\documentclass[sigconf]{acmart}
\AtBeginDocument{%
  }

\setcopyright{cc}
\setcctype{by}
\copyrightyear{2026}
\acmYear{2026}
\acmDOI{10.1145/3767308.3835984}
\acmConference[MM '26]
{34th ACM International Conference on Multimedia}
{November 10--14, 2026}
{Rio de Janeiro, Brazil}

\acmISBN{979-8-4007-2213-4/2026/11}

\usepackage{algorithm}
\usepackage{algpseudocode}
\usepackage{enumitem}
\usepackage{booktabs}
\usepackage[table]{xcolor}
\usepackage{xcolor} 
\definecolor{myblue}{HTML}{D5DFF4} 
\definecolor{upred}{HTML}{BA271A}   
\definecolor{dgreen}{HTML}{1E7D3F}  
\newcommand{\upred}[1]{{\color{upred}$\uparrow$\,#1}}   
\newcommand{\dgreen}[1]{{\color{dgreen}$\downarrow$\,#1}} 
\newcommand{\crossgreen}[1]{{%
  \footnotesize\color{dgreen}#1\,$\times$%
}}
\newcommand{\numbox}[1]{\makebox[6mm][l]{#1}} 
\newcommand{\numboxl}[1]{\makebox[7mm][l]{#1}} 

\begin{document}

\title{Aligning Large Vision–Language Models at Test Time: A Trajectory-Guided Structured Sampling Approach}

\author{Tianbao Jiang}
\orcid{0009-0002-5534-4436}
\affiliation{
  \institution{East China Normal University}
  \city{Shanghai}
  \country{China}
}
\email{tbjiang@stu.ecnu.edu.cn}

\author{Weicong Ni}
\orcid{0009-0009-3399-8827}
\affiliation{
  \institution{East China Normal University}
  \city{Shanghai}
  \country{China}
}
\email{3025394091@qq.com}

\author{Gerard de Melo}
\orcid{0000-0002-2930-2059}
\affiliation{
  \institution{Hasso Plattner Institute}
  \city{Potsdam}
  \country{Germany}
}
\affiliation{
  \institution{University of Potsdam}
  \city{Potsdam}
  \country{Germany}
}
\email{gdm@demelo.org}

\author{Linlin Wang}
\orcid{0000-0003-0304-7560}
\authornote{Corresponding author.}
\affiliation{
  \institution{East China Normal University}
  \city{Shanghai}
  \country{China}
}
\affiliation{
  \institution{City University of Hong Kong}
  \city{Hong Kong}
  \country{China}
}
\email{llwang@cs.ecnu.edu.cn}


\begin{abstract}
Post-training reinforcement learning (RL) algorithms are commonly used to align large vision-language models (LVLMs) with human intent and the requirements of visual reasoning tasks. However, existing RL-based alignment methods are often resource-intensive and encounter mismatches between training objectives and inference-time distributions. To bridge this gap, we propose a novel test-time alignment approach that leverages trajectory-guided structured sampling for dynamic inference-time refinement, achieving better alignment with visual grounding and ensuring logical consistency. Our approach begins with curating a reasoning memory bank via a trajectory learning algorithm, which decomposes complex question solving into ordered sequences of predefined reasoning patterns. It subsequently accomplishes inference-time alignment by first collecting trajectories from reasoning memory bank to establish a global structural reasoning prior, and then using an iterative Markov Chain Monte Carlo (MCMC) algorithm for localized multi-objective refinement of the reasoning trace. Experiments across multiple multimodal reasoning datasets demonstrate that our approach significantly improves accuracy without incurring prohibitive inference overhead. These results establish trajectory-guided test-time sampling as a scalable and effective alternative to traditional post-training alignment, particularly for complex visual reasoning tasks.
\end{abstract}



\maketitle

\section{Introduction}

Despite their impressive performance across many tasks \citep{bai2025qwen25vltechnicalreport, zhu2025internvl3exploringadvancedtraining}, large vision--language models (LVLMs) remain unreliable on complex visual reasoning tasks that require both strong visual grounding and multi-step inference~\citep{zhang-etal-2025-improve, yao2025mulberry}. In such settings, even after post-training, they often fail to sustain visually grounded and logically coherent intermediate reasoning at inference time, drifting from visual evidence and producing unstable reasoning trajectories. Existing alignment approaches, such as reinforcement learning from human feedback (RLHF)~\citep{rlhf-v}, direct preference optimization (DPO)~\cite{liu2025miadpo}, and reinforcement learning with verifiable rewards (RLVR), primarily address this problem through offline parameter updates. However, such training-based methods are computationally expensive and offer limited control over inference-time reasoning, especially under new or unforeseen input distributions.

\begin{figure}[t]
  \includegraphics[width=\columnwidth]{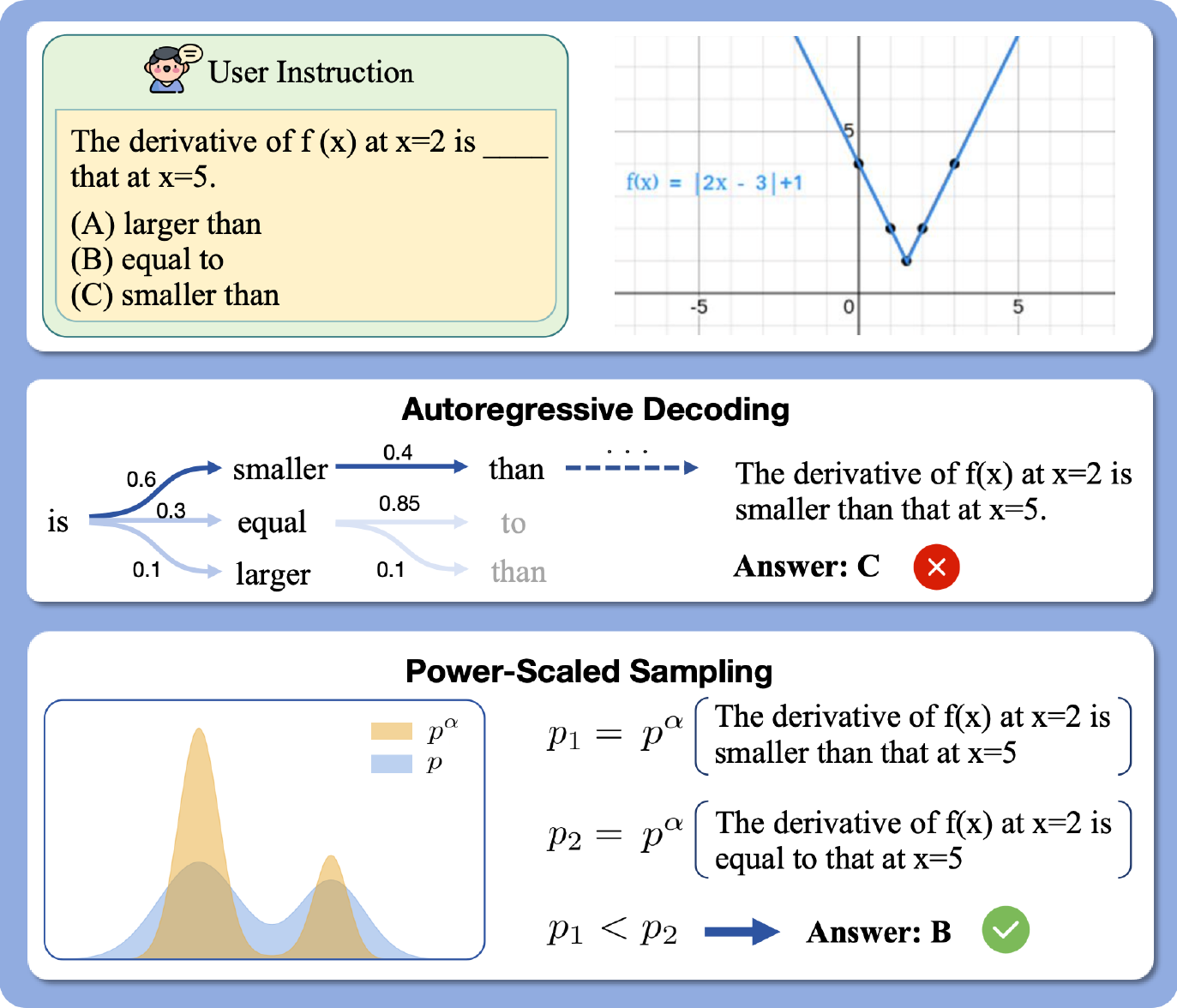}
  \caption{\textbf{Standard autoregressive decoding makes locally optimal choices that may lead to errors. In contrast, power-scaled sampling reweights full sequences via a power-transformed probability $p^\alpha$ ($\alpha>1$).}} 
  \label{fig:core}
\end{figure}

Aligning LVLMs at inference time without modifying their underlying weights emerges as a promising direction for addressing the limitations of training-based alignment methods~\citep{lin2025parm, xu2025genarmrewardguidedgeneration}, such as ensemble-based hypothesis reweighting~\citep{lee2025inferencetime}, and predictive planning-based alignment~\citep{wang2025testtimealignmentlargelanguage}. 
Within this paradigm, sampling-based test-time alignment offers an attractive trade-off by using additional decoding-time compute to search for outputs that better satisfy task objectives under distribution shift. In practice, this is often achieved by reshaping the model's output distribution, for example through reward model guided reweighting~\citep{lin2025parm}, importance-weighted perturbations~\citep{kanai2025testtimealignmentllmssamplingbased}, or iterative refinement. 
More recently, \citet{karan2025reasoningsamplingbasemodel} demonstrate that sampling from the power-scaled distribution $p^\alpha$ via Metropolis--Hastings (MH) yields training-free reasoning improvements that are comparable to those obtained with RLVR. 
As shown in Figure~\ref{fig:core}, sequence-level reweighting in power-scaled sampling avoids misleading local token choices and better preserves visual evidence than autoregressive decoding.

Despite its promise, applying MH-based power sampling to LVLMs for visual reasoning remains challenging. First, power scaling may amplify language priors~\citep{wang2025mllm} and visually unsupported details when visual evidence is weak, exacerbating hallucinations. 
Second, complex visual reasoning requires long-horizon, multi-step deduction with multiple plausible solution paths. For example, solving the question in Figure~\ref{fig:core} requires the model to identify the derivatives at both $x=2$ and $x=5$ before comparing them. 
Sampling can easily drift into high-entropy, unstable reasoning
branches.
Finally, for long-horizon generation, the induced Markov chain often mixes slowly in a high-dimensional token space~\citep{brown2025upperlowerboundssubgeometric}, 
resulting in low acceptance and many expensive proposal rollouts, which substantially increases inference latency.

To address these limitations, we propose a training-free test-time alignment framework that allocates computation to targeted, structured refinement during decoding. We first build a reasoning memory bank through our trajectory learning algorithm, where each trajectory encodes the solution process for a complex question as an ordered sequence of high-level reasoning patterns. Given a query, we retrieve the top-$k$ most similar examples from the bank and aggregate their trajectories via majority voting to obtain a single guidance trajectory. This trajectory provides a global prior over the reasoning structure and step order, while the LVLM is responsible for instantiating each pattern into task-specific reasoning content. Based on this decomposition, our structured sampling algorithm initializes one reasoning pattern per iteration and performs MCMC refinement within a sliding window of neighboring reasoning patterns. This localized refinement yields low-variance proposals and mitigates slow mixing, making MCMC practical for long-form LVLM generation. On top of this sampler, we define three complementary objectives to shape the power-scaled target distribution: vision-aware distribution sharpening to facilitate visual grounding, an entropy regularizer to favor reliable trajectories, and a linguistic control term to suppress degenerative generation.
In summary, the integration of trajectory-guided structured proposals, iterative Metropolis--Hastings inference, and our synergistic objectives enables effective test-time alignment for complex visual reasoning. Our main contributions are three-folds:

\begin{itemize}
    \item We propose an automatic trajectory learning algorithm rooted in an agentic framework, designed to construct and store candidate trajectories, thereby building a continuous memory of reasoning processes.
    \item We develop a trajectory-guided structured sampling method that retrieves trajectories from memory as a structural reasoning prior, followed by iterative MCMC trajectory refinement that promotes visual grounding and stable generation through vision-aware distribution sharpening, entropy regularization, and linguistic degeneration control, thereby ensuring alignment during inference.
    \item Extensive experiments across multiple reasoning benchmarks demonstrate consistent gains in both accuracy and sampling efficiency, establishing our training-free test-time sampling approach as a highly effective alternative to RLVR.
\end{itemize}

\section{Preliminaries}

\label{sec:power}

\textbf{Power-Scaled Sampling}. Let $V$ and $X$ denote the visual and textual inputs, respectively, and let $\mathbf{Y}=\langle y_0,\ldots,y_T\rangle$ be a finite output token sequence with $y_t\in\mathcal{V}$, where $\mathcal{V}$ is the model vocabulary. An LVLM induces a normalized autoregressive distribution over $\mathcal{Y}$:
\begin{equation}
\label{eq:ar_seq}
P(\mathbf{Y}\mid V,X)=\prod_{t=0}^{T} P(y_t\mid V,X,y_{<t}),
\end{equation}
where $\mathcal{Y}$ denotes the set of all finite sequences. Recent work shows that strong reasoning gains can be obtained by sampling from a power-transformed sequence distribution~\citep{karan2025reasoningsamplingbasemodel}. Given an exponent $\alpha>1$, the corresponding unnormalized target is
\begin{equation}
\label{eq:dis_power}
p_{\alpha}(\mathbf{Y}\mid V,X)\ \propto\ P(\mathbf{Y}\mid V,X)^{\alpha},
\end{equation}
which amplifies differences in sequence likelihood and concentrates probability mass on high-likelihood trajectories under the base model. 
Power sampling operates at the sequence level by reweighting complete continuations $\mathbf{Y}$, rather than making greedy or temperature-scaled decisions at each step, as illustrated in Figure~\ref{fig:core}. Sampling from the power distribution implicitly favors longer-horizon planning by prioritizing globally coherent, high-likelihood trajectories and mitigating failures induced by \textit{pivotal tokens}~\citep{abdin2024phi4technicalreport} that steer generation toward low-likelihood continuations.

\begin{figure*}[t]
  \includegraphics[width=\textwidth]{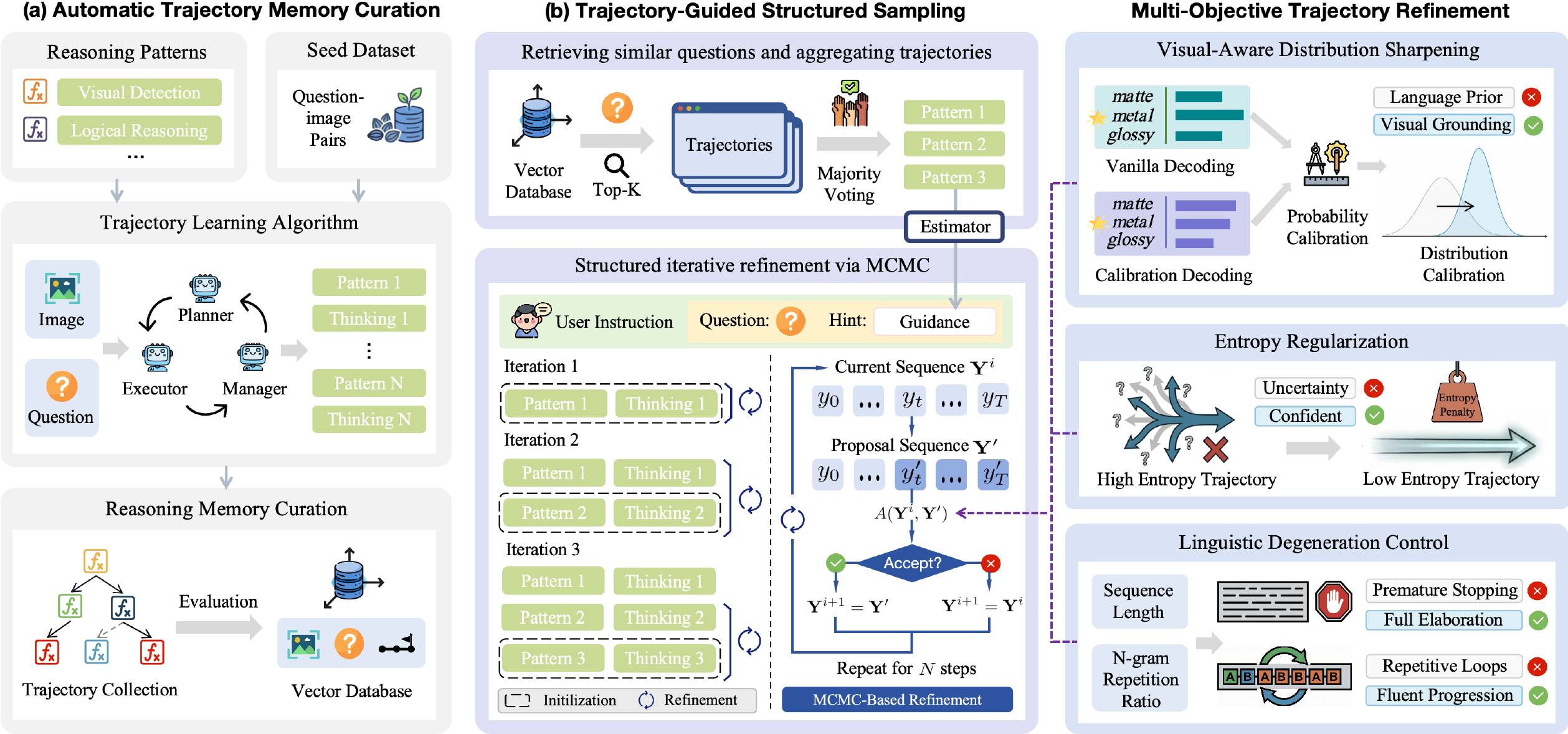}
    \caption{\textbf{Overview of our approach.} We begin by constructing a reasoning memory bank based on a trajectory learning algorithm. At inference time, retrieved trajectories provide a global structural reasoning prior, while MCMC performs localized updates within selected reasoning segments to iteratively steer generation toward the alignment target.}
  \label{fig:framework}
\end{figure*}

\section{Methodology}

In this section, we present our test-time alignment approach for visual reasoning with LVLMs (Figure~\ref{fig:framework}). We first introduce the automatic trajectory memory curation pipeline (Section~\ref{sec:library}), which uses an agentic framework to search for effective reasoning paths under predefined reasoning patterns. We then describe the structured sampling algorithm (Section~\ref{sec:trajec-samp}), where trajectory guidance provides a structural prior for LVLM reasoning and Metropolis--Hastings is used to perform iterative refinement within a local window of adjacent reasoning patterns. Finally, we introduce our refinement target (Section~\ref{sec:alignment_obj}), which combines vision-aware distribution sharpening, entropy regularization, and linguistic degeneration control.

\label{sec:library}
\begin{figure*}[t]
  \includegraphics[width=\textwidth]{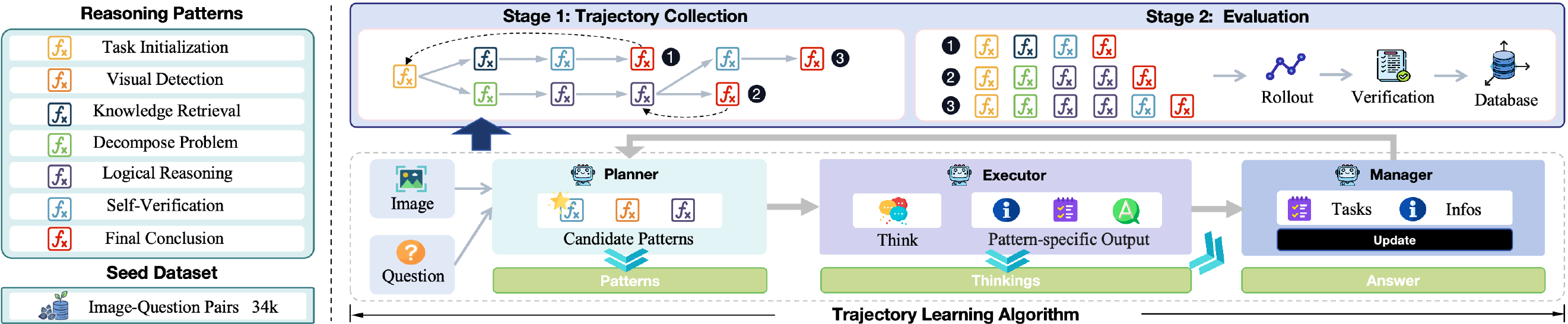}
  \caption{\textbf{Automatic trajectory memory curation.} After defining reasoning patterns and preparing the seed dataset, the reasoning memory bank is built through a two-stage reasoning memory curation process.}
  \label{fig:pipeline}
\end{figure*}

\subsection{Automatic Trajectory Memory Curation}
\label{sec:library}

Complex multimodal questions often require high-level reasoning
behaviors, such as decomposition~\citep{li2025selfrewardingvisionlanguagemodelreasoning}, knowledge grounding, and self-verification~\citep{wang2025vlrethinker}. However, standard LVLM decoding may fail to activate these behaviors reliably, leading to brittle trajectories on challenging instances. We therefore introduce an automated trajectory discovery pipeline that decomposes the reasoning process into discrete patterns, curating a reusable memory bank of structured reasoning traces to facilitate test-time guidance (Figure~\ref{fig:pipeline}).

Based on insights from prior works ~\citep{wu2026astarboostingmultimodalreasoning,li2025thinkpilotsteeringreasoningmodels}, we define seven reasoning patterns, as shown in Figure~\ref{fig:pipeline}. Rather than relying on heuristic prompts or taxonomies of reasoning actions~\cite{li-etal-2025-understanding}, we describe each pattern in a pseudo-code style
, making offline reasoning experience explicit and reusable during inference. We construct the seed set using 34k de-duplicated question--image pairs from ThinkLite-70k~\citep{wang2025sotalessmctsguidedsample}, covering mathematical reasoning, natural image understanding, and chart understanding.

\noindent\textbf{Trajectory Learning Algorithm.}
Motivated by~\citet{lu2025octotoolsagenticframeworkextensible}, we develop an agentic framework that automatically learns trajectories through iterative collaboration among a planner, an executor, and a manager, as shown in Figure~\ref{fig:pipeline}. A standard chain-of-thought (CoT) can be represented as
$\mathbf{Y}=\{r_0,r_1,\dots,r_T\}$,
where $r_t$ denotes the $t$-th rationale and $r_T$ contains the final answer. To make a reasoning trajectory executable and editable at the step level, we represent each intermediate state using (i) a task list $\mathbf{T}$, which specifies the remaining tasks, and (ii) an information list $\mathbf{I}$, which stores the intermediate results and supporting evidence collected so far. This state representation also provides explicit inputs and outputs for each reasoning pattern. Accordingly, the CoT is reformulated as:
\begin{equation}
\mathbf{Y} =
\{(\mathbf{T}_{0},\mathbf{I}_{0}),
(\mathbf{T}_{1},\mathbf{I}_{1}),
\dots,
(\mathbf{T}_{T},\mathbf{I}_{T})\},
\end{equation}
where $\mathbf{I}_{T}$ contains the final answer.
Given an image $V$ and a question $X$, the planner LVLM $M_{\theta}$ selects the next reasoning pattern $a_t$ at step $t$. Its decision is conditioned on the first pending task $\mathbf{T}_t^{(0)}$ and the reasoning history $H_t$, which records the previously selected patterns and their associated thinking contexts:
\begin{equation}
a_t = M_{\theta}(V, X, \mathbf{T}_t^{(0)}, H_t),
\quad t \ge 0.
\end{equation}
To facilitate exploration, we rank patterns by their first-token probabilities and sample from the top candidates. The executor LVLM then applies the selected pattern to generate an intermediate thinking context $c_t$ and the corresponding pattern-specific outputs $o_t$:
\begin{equation}
c_t, o_t =
M_{\theta}(V, X, a_{\le t}, \mathbf{T}_t, \mathbf{I}_t).
\end{equation}
Finally, the manager LVLM uses $c_t$ and $o_t$ to transform the current state $(\mathbf{T}_t,\mathbf{I}_t)$ into the next state $(\mathbf{T}_{t+1},\mathbf{I}_{t+1})$. Specifically, it removes completed tasks from $\mathbf{T}_t$ and updates $\mathbf{I}_t$ by adding newly obtained information or revising existing entries. The updated lists then provide a clear starting state for the next reasoning step.

\noindent\textbf{Reasoning Memory Curation.}
We use depth-first search to collect candidate trajectories. At each step, we retain the highest-ranked candidate patterns until their cumulative probability reaches a threshold $\tau\in(0,1)$, and backtrack when a terminal pattern is reached. To estimate reliability, we roll out each candidate trajectory $k$ times and calculate its answer correctness rate, retaining the most reliable traces. For downstream retrieval, we index the question embeddings in a vector database and store the best-performing trajectory for each question as retrievable metadata.

\subsection{Trajectory-Guided Structured Sampling}

At inference, we retrieve the top-$k$ questions most similar to input $(V,X)$. Candidates are ranked by an entropy-weighted score combining image and text embedding similarities with $n$-gram overlap, where lower-entropy views receive larger weights. The associated trajectories form:
$
\mathcal{T}(V,X) = \{\boldsymbol{\tau}^{(j)}\}_{j=1}^{k},
$
where each trajectory is an ordered sequence of predefined reasoning patterns. 
Given the retrieved set, we construct a guidance trajectory by majority voting at each aligned position:
\begin{equation}
\boldsymbol{\tau}^{*}=\langle a^{*}_1,\ldots,a^{*}_{L}\rangle.
\end{equation}
Starting from all retrieved trajectories, at each step $t$, we select the most frequent pattern as $a_t^{*}$ and retain only trajectories matching the selected prefix $\langle a_1^{*},\ldots,a_t^{*}\rangle$ for the next step.

Metropolis--Hastings~\citep{Metropolis1953Equation} is a classic Markov Chain Monte Carlo (MCMC) algorithm that approximately samples from an unnormalized target distribution via iterative proposals and a stochastic accept--reject rule.
Let $\mathbf{Y}^0_{0:T}$ denote the current sequence being refined, and let $p_{\text{prop}}$ be the proposal LVLM. At iteration $i$, we sample a boundary index $t \sim \mathcal{U}\{0,\dots,T\}$ and propose $\mathbf{Y}'$ by resampling the suffix $\mathbf{Y}^i_{t:T}$ with $p_{\text{prop}}$ while keeping the prefix $\mathbf{Y}^i_{0:t-1}$ fixed. The proposal is
accepted with probability
\begin{equation}
\label{eq:r}
A(\mathbf{Y}^i,\mathbf{Y}')=\min\left\{1,\frac{p_{t}(\mathbf{Y}')\,q(\mathbf{Y}^i\mid \mathbf{Y}')}{p_{t}(\mathbf{Y}^i)\,q(\mathbf{Y}'\mid \mathbf{Y}^i)}\right\},
\end{equation}
where $p_{t}$ is the unnormalized target density over sequences and $q$ is the
proposal transition. Repeating for $N$ iterations yields a chain
$\{\mathbf{Y}^0,\ldots,\mathbf{Y}^N\}$ that approximately samples from the target.

Given the trajectory guidance, we first use a rule-based estimator that takes the aggregated trajectory as input to route trivial questions to full-sequence MCMC refinement in order to reduce overthinking and improve efficiency.
For non-trivial questions, we prompt the LVLM to reason by following the guided trajectory and do iterative sampling as shown in Algorithm~\ref{alg:vspar_simple}.
We highlight three design choices that are central to our approach.
\label{sec:trajec-samp}
\begin{algorithm}[t]
\caption{\textsc{Trajectory-Guided Structured Sampling}}
\label{alg:vspar_simple}
\begin{algorithmic}[1]
\Require proposal LVLM $p_{\text{prop}}$, inputs $(V,X)$, max length $T$, initial temperature $\tau_0$, initial sharpening exponent $\alpha_0$, decay threshold $\kappa$, decay factor $\gamma$, MCMC steps $N_{\text{MCMC}}$
\State $\mathcal{B}\gets[\,]$ \Comment{Initialize block boundaries; $\mathcal{B}$ stores start indices of reasoning blocks}
\State $\tau \gets \tau_0$, $\alpha \gets \alpha_0$
\While{$|\mathbf{Y}| < T$}
  \State \textbf{Annealed Sampling Schedule:} 
  \State \hspace{1em} $\tau \gets \text{max}(0.1,\,\tau \cdot \gamma^{\mathbb{I}[|\mathcal{B}|>\kappa]}$), $\alpha \gets \tau^{-1}$

  \State \textbf{Pattern-wise Initialization:} 
  \State \hspace{1em} $U \gets |\mathcal{B}| + 1$ \Comment{the max numbers of pattern to generate}
  \State \hspace{1em} $\mathbf{Y} \gets \Call{Generation}{p_{\text{prop}},\,\mathbf{Y},\,U,\tau,\,\alpha}$

  \State \textbf{Sliding-Window Sampling:} perform MCMC-based refinement within the newest two reasoning patterns.
  \State \hspace{1em} $\mathbf{Y}\gets \Call{MCMC}{p_{\text{prop}},\mathbf{Y},N_{\text{MCMC}},\,\tau\,\alpha}$

  \State Update $\mathcal{B}$ according to updated sequence $\mathbf{Y}$
  \State \textbf{if} \Call{HasTerminal}{$\mathbf{Y}$} \textbf{then break}
\EndWhile
\State \Return $\mathbf{Y}$
\end{algorithmic}
\end{algorithm}

\noindent\textbf{(i) Pattern-wise Initialization}. Each iteration adds at most one new reasoning pattern, thus limiting proposal drift and improving acceptance in high-dimensional sequence spaces~\citep{brown2025upperlowerboundssubgeometric}. This pattern-level granularity also makes the refinement process interpretable.

\noindent\textbf{(ii) Sliding-Window Sampling}. Full-sequence resampling is costly for long outputs. Instead, we update only a local suffix over the most recent two reasoning patterns, reducing inference cost while retaining the benefits of sequence-level refinement.

\noindent\textbf{(iii) Annealed Sampling Schedule}. Since windowed MH refines recent patterns on top of a mostly fixed prefix, excessive temperature can destabilize the existing context. We therefore decrease the sampling temperature $\tau$ and increase the sharpening exponent $\alpha$ only when the number of reasoning patterns $|\mathcal{P}|$ exceeds a threshold $\kappa$, yielding a smooth transition from exploration to exploitation.



\subsection{Multi-Objective Trajectory Refinement}
\label{sec:alignment_obj}

Trajectory guidance provides a global prior over desirable reasoning patterns, but local realizations can still drift from visual evidence or fall into decoding pathologies. In this section, we define our test-time alignment target that augments power-based sharpening with step-level alignment objectives, so that alignment can be improved via MCMC refinement without updating model parameters.

\subsubsection{Vision-Aware Distribution Sharpening}
\label{sec:vision_aware}
Naively sharpening the sequence distribution can overemphasize language priors under weak visual evidence, leading to increased hallucinations. Inspired by contrastive visual decoding~\citep{VCD,clearsight}, we introduce a vision-aware calibration mechanism that reweights next-token probabilities against a ungrounded baseline. Specifically, the ungrounded branch is constructed by masking visual tokens during prefill, while subsequent decoding steps are synchronized with the visual-grounded branch through teacher-forcing. Let $y_t$ be the token chosen by the grounded branch at step $t$, we thereby define the alignment ratio $r_t$ as:
\begin{equation}
\label{eq:rt}
 r_t = \frac{P_{\text{visual-grounded}}(y_t) + \delta}{P_{\text{ungrounded}}(y_t) + \delta},
\end{equation}
where $\delta$ is a small smoothing constant for numerical stability. Intuitively, a larger $r_t$ suggests that the current token is more visually grounded. We then convert $r_t$ into a bounded calibration weight $s(r_t)$, following adaptive calibration~\citep{huo2025selfintrospective,evrt}:
\begin{equation}
\label{eq:sr}
s(r_t)=
\begin{cases}
\exp\!\Big(-\beta \big[\operatorname{softplus}(-\ln r_t-\epsilon)-\operatorname{softplus}(-\epsilon)\big]\Big), & r_t<1,\\
1, & r_t\ge 1,
\end{cases}
\end{equation}
where $\beta$ controls suppression strength, $\epsilon=10^{-3}$ is a small margin, and $\operatorname{softplus}(x)=\ln(1+e^{x})$. The subtraction term makes the penalty zero at the boundary, ensuring continuity as $r_t\to 1^{-}$.
Next-token probabilities are calibrated with $s(r_t)$ as
\begin{equation}
\label{eq:tildeP_token}
\tilde{P}(y_t \mid V,X,y_{<t}) \;=\; P(y_t \mid V,X,y_{<t}) \cdot s(r_t),
\end{equation}
resulting in a vision-aware sharpened sequence likelihood:
\begin{equation}
\tilde{P}(\mathbf{Y}\mid V,X)^{\alpha} \;=\;(\prod_{t=0}^{T}\tilde{P}(y_t\mid V,X,y_{<t}))^{\alpha}.
\end{equation}



\subsubsection{Entropy Regularization}
\label{sec:entropy_reg}
Recent evidence suggests that RL-style alignment is often accompanied by reduced predictive entropy~\citep{cui2025entropy}. Consistently, our preliminary experiment shows that correct generations tend to exhibit lower predictive entropy under the base model. We therefore exponentially downweight trajectories according to their cumulative token-level predictive entropy:
\begin{equation}
\label{eq:entropy_penalty}
R_H(\mathbf{Y})=
\exp\!\left(-\lambda_H \sum_{t=0}^{T} H\!\left[P(\cdot \mid V,X,y_{<t})\right]\right),
\end{equation}
where $\lambda_H$ controls the strength of the entropy penalty.

\subsubsection{Linguistic Degeneration Control}
\label{sec:length_ctrl}

To reduce premature stopping while discouraging repetitive loops, we introduce a linguistic control term that couples effective length with an explicit repetition signal. Let $\ell(\mathbf{Y})$ be the sequence length and let $\rho(\mathbf{Y})\in[0,1]$ denote an $n$-gram repetition ratio. We define the effective length score as:
\begin{equation}
\label{eq:eff_len}
S_L(\mathbf{Y}) \;=\; \frac{\ell(\mathbf{Y})}{1 + \ln(\ell(\mathbf{Y}))\,\rho(\mathbf{Y})},
\end{equation}
which grows with length but is increasingly discounted as repetition increases, especially for long outputs through the $\ln(\ell(\mathbf{Y}))$ factor. We incorporate a linguistic degeneration control factor by exponentiating a shaped effective-length reward:
\begin{equation}
\label{eq:length_factor}
R_L(\mathbf{Y}) \;=\;
\exp\!\left(L\cdot\Big(1-\exp\big(-a\cdot S_L(\mathbf{Y})^{\,b}\big)\Big)\right),
\end{equation}
where $L$ sets the saturation level, $a$ controls the sensitivity to changes in $S_L(\mathbf{Y})$, and $b$ shapes the degree of diminishing returns. 
\subsubsection{Synergistic Joint Alignment Target}
\label{sec:static_target_mcmc}

By integrating these objectives, we define an unnormalized alignment target over complete sequences, with $\alpha>1$ controls the sharpening strength:
\begin{equation}
\label{eq:aligned_target}
p_t(\mathbf{Y}\mid V,X)\ \propto\ \tilde{P}(\mathbf{Y}\mid V,X)^{\alpha}\cdot R_H(\mathbf{Y})\cdot R_L(\mathbf{Y}).
\end{equation}
This synergistic formulation yields a balanced energy landscape. The vision-aware term first filters out visually unsupported reasoning branches, after which entropy regularization biases MCMC toward reliable, confident trajectories. Crucially, while entropy regularization alone may favor confidently wrong outputs, its coupling with visual calibration ensures that only visually consistent paths are sharpened. Finally, the linguistic control term prevents collapse into repetitive loops or premature termination. Together, these objectives turn test-time sampling into a robust optimization process that yields concise and logically coherent reasoning trajectories.

\section{Experiments}
\begin{table*}[t]
\centering
\caption{\textbf{Performance comparison on five datasets.} }
\label{tab:main_results}
\small
\begin{tabular}{p{38mm}|p{19mm}|p{19mm}|p{19mm}|p{19mm}|p{19mm}|p{19mm}}
\toprule
Method & MathVista & MathVision & MathVerse & MMMU & MMStar & Avg. \\ 
\midrule
\multicolumn{7}{c}{\textit{General Vision-Language Models}} \\ 
\midrule
InternVL2.5-8B~\citep{chen2025expandingperformanceboundariesopensource} & 64.4 & 22.0 & 39.5 & 54.9 & 62.8 & 48.7  \\
Qwen2.5-VL-7B~\citep{bai2025qwen25vltechnicalreport} & 68.2 & - & 49.2 & 58.6 & 63.9 & -  \\
InternVL3-8B~\citep{zhu2025internvl3exploringadvancedtraining} & 71.6  & 29.3 & 46.3 & 62.7 & 68.2 & 55.6  \\
Qwen3-VL-8B~\citep{Qwen3-VL} & 77.2 & - & 62.1 & 69.6 & 70.9 & -  \\

\midrule
\multicolumn{7}{c}{\textit{RLVR-Trained Vision-Language Models}} \\ 
\midrule
OpenVLThinker-7B~\citep{deng2025openvlthinkercomplexvisionlanguagereasoning} & 70.2  & 29.6 & 50.3 & 51.9 & 63.2 & 53.0  \\
VL-Rethinker-7B~\citep{wang2025vlrethinker} & 74.9 & 32.3 & 54.2 & 56.7 & - & -  \\
Vision-R1-7B~\citep{huang2025visionr1incentivizingreasoningcapability} & 73.5  & 30.7 & 52.4 & 50.5 & 60.2 & 53.4  \\
MM-EUREKA-7B~\citep{meng2025mmeurekaexploringfrontiersmultimodal} & 73.0  &  31.9 & 50.3 & 52.3 & 64.1 & 54.3  \\
ThinkLite-VL-7B~\citep{wang2025sotalessmctsguidedsample} & 75.1  & 32.9 & 52.1 & 55.5 & 65.0 & 56.1  \\
\midrule
\multicolumn{7}{c}{\textit{Sampling-based Test Time Alingment (Qwen2.5-VL-7B)}} \\ 
\midrule
Qwen2.5-VL-7B (Reproduced) & 70.0 & 27.3 & 47.1 & 54.0 & 61.3 & 52.0  \\
\rowcolor{myblue}
+ Multi-Objective Sampling 
  & \numbox{\textbf{73.2}}\;\upred{3.2} 
  & \numbox{\textbf{30.9}}\;\upred{3.6} 
  & \numbox{\textbf{50.0}}\;\upred{2.9} 
  & \numbox{\textbf{54.9}}\;\upred{0.9} 
  & \numbox{64.0}\;\upred{2.7} 
  & \numbox{\textbf{54.6}}\;\upred{2.6} \\
  
\rowcolor{myblue}
+ TG Structured Sampling 
  & \numbox{72.7}\;\upred{2.7} 
  & \numbox{30.3}\;\upred{3.0}
  & \numbox{49.5}\;\upred{2.4}
  & \numbox{54.7}\;\upred{0.7}
  & \numbox{\textbf{64.5}}\;\upred{3.2} 
  & \numbox{54.3}\;\upred{2.3}  \\
\bottomrule
\end{tabular}
\end{table*}
\subsection{Experiment Settings}

\subsubsection{Datasets}
We evaluate our framework on diverse multimodal benchmarks requiring multi-hop reasoning, allowing us to assess test-time alignment under long-horizon inference. We cover multimodal math reasoning through MathVista~\citep{lu2024mathvista}, MathVerse~\citep{zhang2024mathverse}, and MathVision~\citep{wang2024measuring}, while targeting college-level multi-disciplinary tasks in MMMU~\citep{yue2023mmmu} and general LVLM robustness in MMStar~\citep{chen2024are}.





\subsubsection{Baselines}
We compare against two categories of models:
\begin{itemize}[leftmargin=*,itemsep=0.2em]
    \item \textbf{Strong general-purpose LVLMs.}
    We include widely used open-source backbones with strong general multimodal performance across model generations: InternVL2.5-8B~\citep{chen2025expandingperformanceboundariesopensource}, Qwen2.5-VL-7B~\citep{bai2025qwen25vltechnicalreport}, InternVL3-8B~\citep{zhu2025internvl3exploringadvancedtraining}, and Qwen3-VL-8B~\citep{Qwen3-VL}.

    \item \textbf{RLVR-trained LVLMs}
    We include representative reasoning-oriented LVLMs that improve multi-step inference through RL with verifiable rewards (RLVR),such as OpenVLThinker-7B~\citep{deng2025openvlthinkercomplexvisionlanguagereasoning}, VL-Rethinker-7B~\citep{wang2025vlrethinker}, Vision-R1-7B~\citep{huang2025visionr1incentivizingreasoningcapability}, MM-EUREKA-7B~\citep{meng2025mmeurekaexploringfrontiersmultimodal}, ThinkLite-VL-7B~\citep{wang2025sotalessmctsguidedsample}. All the baselines share the same Qwen2.5-VL-7B backbone for fair comparison.
\end{itemize}

\subsubsection{Implementation details} For trajectory discovery pipeline, we adopt Qwen3-VL-8B as the planner, executor, and manager.
After verification, we retain 24k question--trajectory pairs and index the
questions in a vector database. In our main experiments, we report results for two variants: (i) \textbf{Multi-Objective Sampling} without trajectory guidance, where we prompt the model to produce intermediate reasoning before answering and perform full-sequence MCMC refinement, as in~\citet{karan2025reasoningsamplingbasemodel}. and (ii) \textbf{Trajectory-Guided (TG) Structured Sampling}, which further incorporates trajectory guidance and block-wise refinement for non-trivial questions. We set the sharpening exponent $\alpha=4$ and the generation temperature $\tau=0.25$. For the alignment objectives, we use visual calibration weight $\beta=2.0$, entropy weight $\lambda_H=1.0$. For linguistic control term, we set $a=0.05$, $b=0.8$ and linguistic control saturation level $L=128$. For TG structured sampling, we set the decay threshold $\kappa=4$ and decay factor $\gamma=0.5$. We evaluate our methods on Qwen2.5-VL-7B, a widely used LVLM that has shown strong potential in prior RL-based studies.

\begin{table}[t]
\centering
\small
\caption{\textbf{Ablation on trajectory-guided structured sampling.}}
\label{tab:ablation_tgss}
\begin{tabular}{p{31mm}|p{13mm}|p{13mm}|p{13mm}}
\toprule
Method & MathVista & MMStar & Mathvision \\
\midrule
\rowcolor{myblue}
TG Structured Sampling & \textbf{72.7} & \textbf{64.5} & \textbf{30.3} \\
w/o Sampling Schedule & \numbox{71.9}\dgreen{0.8} & \numbox{63.6}\dgreen{0.9} & \numbox{28.6}\dgreen{1.7} \\
w/o Trajectory Guidance & \numbox{71.8}\dgreen{0.9} & \numbox{63.8}\dgreen{0.7} & \numbox{28.0}\dgreen{2.3} \\
\bottomrule
\end{tabular}
\end{table}

\begin{table*}[t]
\centering
\caption{\textbf{Ablation study on alignment target. We sample directly from Qwen2.5-VL-7B under different MCMC target distributions.}}
 \label{tab:sampling_objs}
\small
\begin{tabular}{p{62mm}|p{15mm}|p{15mm}|p{15mm}|p{15mm}|p{15mm}|p{15mm}}
\toprule
Method & MathVista & MathVision & MathVerse & MMMU & MMStar & Average \\
\midrule
Qwen2.5-VL-7B (Reproduced) & 70.0 & 27.3 & 47.1 & 54.0 & 61.3 & 52.0 \\
+ Power Sampling & 72.0 & 29.9 & 49.3 & 52.7 & 63.1 & 53.4  \\
+ Power Sampling + Visual & 70.8 & 30.9 & 49.5 & 53.6 & 64.1 & 53.8  \\
+ Power Sampling + Entropy & 70.2 & 25.0 & 46.6 & 52.1 & 62.7 &  51.3 \\
+ Power Sampling + Entropy + Visual & 72.2 & \underline{\textbf{31.3}} & 49.8 & 54.7 &  \underline{\textbf{64.4}} & 54.5 \\
+ Power Sampling + Entropy + Visual + Linguistic &  \underline{\textbf{73.2}} &  30.9 &  \underline{\textbf{50.0}} &  \underline{\textbf{54.9}} & 64.0 & \underline{\textbf{54.6}} \\
\bottomrule
\end{tabular}
\end{table*}

\begin{table}[t]

\centering
\small
\caption{\textbf{Performance on MMStar and MathVision with InternVL2.5-8B and Qwen3-VL-8B.}}
\label{tab:general}
\begin{tabular}{p{40mm}|p{17mm}|p{17mm}}
\toprule
Method & MMStar & MathVision  \\
\midrule

InternVL2.5-8B~\citep{chen2025expandingperformanceboundariesopensource} & 61.2 & 21.4  \\
\rowcolor{myblue}
+ Multi-Objective Sampling  & \numbox{\textbf{62.8}}\;\upred{1.6} & \numbox{\textbf{25.0}}\;\upred{3.6}  \\
Qwen3-VL-8B~\citep{Qwen3-VL} & 65.3 & 47.7  \\
\rowcolor{myblue}
+ Multi-Objective Sampling & \numbox{\textbf{73.2}}\;\upred{7.9} & \numbox{\textbf{51.0}}\;\upred{3.3}  \\
\bottomrule
\end{tabular}
\end{table}

\subsection{Main Results}
\label{sec:main_results}

Table~\ref{tab:main_results} reports results on five challenging multimodal datasets. Overall, our methods consistently improves the reproduced Qwen2.5-VL-7B baseline across all datasets. Multi-Objective Sampling yields the best average accuracy ($54.6$), while TG Structured Sampling remains highly competitive ($54.3$). Three takeaways stand out.

Training-free test-time alignment is competitive with RLVR-trained reasoning models.
Without any post-training, our sampling-based refinement closes much of the gap to RLVR-aligned LVLMs and outperforms several RLVR-trained baselines on the reported suite, suggesting that reallocating compute to inference-time refinement is an effective way to improve alignment. 

Trajectory-guided structured sampling preserves accuracy and improves efficiency.
Compared to full-sequence refinement, TG Structured Sampling achieves similar accuracy while using step-aligned, block-wise proposals guided by a voted trajectory prior, reducing unnecessary resampling and making long-horizon refinement more practical (efficiency results in Section~\ref{sec:efficiency}).

General-purpose trajectory priors help most on general dataset.
Since the trajectory memory is built from a general-purpose seed set (ThinkLite-70k), trajectory guidance is particularly effective on the general dataset MMStar. On more reasoning-intensive math datasets, gains remain consistent but can be slightly smaller, suggesting room for more domain-specialized trajectory libraries.

\subsection{Ablation Study on Structured Sampling}
\label{sec:ablation_tgss}

Table~\ref{tab:ablation_tgss} ablates two key components of trajectory-guided structured sampling.
Removing annealed sampling schedule consistently degrades performance across all evaluated benchmarks, indicating that a gradual shift from exploration to exploitation is important for stable long-horizon refinement.
Disabling trajectory guidance leads to a larger drop, especially on MathVision, suggesting that the retrieved-and-voted trajectory prior provides useful global structure that improves the quality of block-level proposals and subsequent MCMC refinement.
Notably, this sensitivity to temperature is amplified by our relatively strong alignment target: overly aggressive exploration can cause block-wise updates to overfit local edits that score well under the target while drifting from a globally coherent solution, trapping the chain in suboptimal regions and reducing final accuracy.
Overall, both components contribute to the final accuracy, with trajectory guidance being the primary driver and annealed sampling schedule offering complementary stabilization.

\subsection{Ablation Study on Alignment Target}
\label{sec:ablation_obj}

Table~\ref{tab:sampling_objs} ablates our test-time alignment target (Eq.~\ref{eq:aligned_target}). Power sampling improves upon the reproduced Qwen2.5-VL-7B baseline across several datasets, indicating that sharpening the sequence distribution is a strong training-free lever. Incorporating the vision-aware term yields further gains over power sampling on every dataset except MathVista, aligning with its role in suppressing language-prior amplification under weak or ambiguous visual evidence.

Adding entropy regularization alone consistently underperforms power sampling, suggesting a failure mode in which low-entropy trajectories are favored even when they encode confident mistakes, thereby limiting error correction. Coupling it with the vision-aware term alleviates this issue by filtering unsupported tokens and stabilizing visually grounded refinement. Among all ablations, the joint objective closely matches the best setting.

Adding the linguistic control term yields further gains, most notably on MathVista. Results on MMStar (typically short) and MathVision (often long) suggest that a fixed length prior is not universally optimal, and the parameters $L$ should instead be calibrated to the length regime of the target dataset.

\subsection{Generalization of Alignment Target}

To further assess the generalizability of our refinement target, we evaluate it on two additional LVLMs, InternVL2.5-8B and Qwen3-VL-8B, with results summarized in Table~\ref{tab:general}. We use the same core experimental settings as for Qwen2.5-VL-7B to test whether the refinement target transfers across model families. In particular, both the visual calibration weight ($\beta = 2.0$) and the entropy weight ($\lambda_H = 1.0$) remain effective without retuning. The only parameter that requires mild model-specific adjustment is the linguistic control saturation level $L$, likely due to differences in linguistic priors and response verbosity. Even so, our method produces consistent improvements across base architectures. Overall, these findings indicate that the proposed synergistic alignment target generalizes well across models and serves as a practical plug-and-play component for improving LVLM reasoning.

\begin{figure}[t]
  \includegraphics[width=\columnwidth]{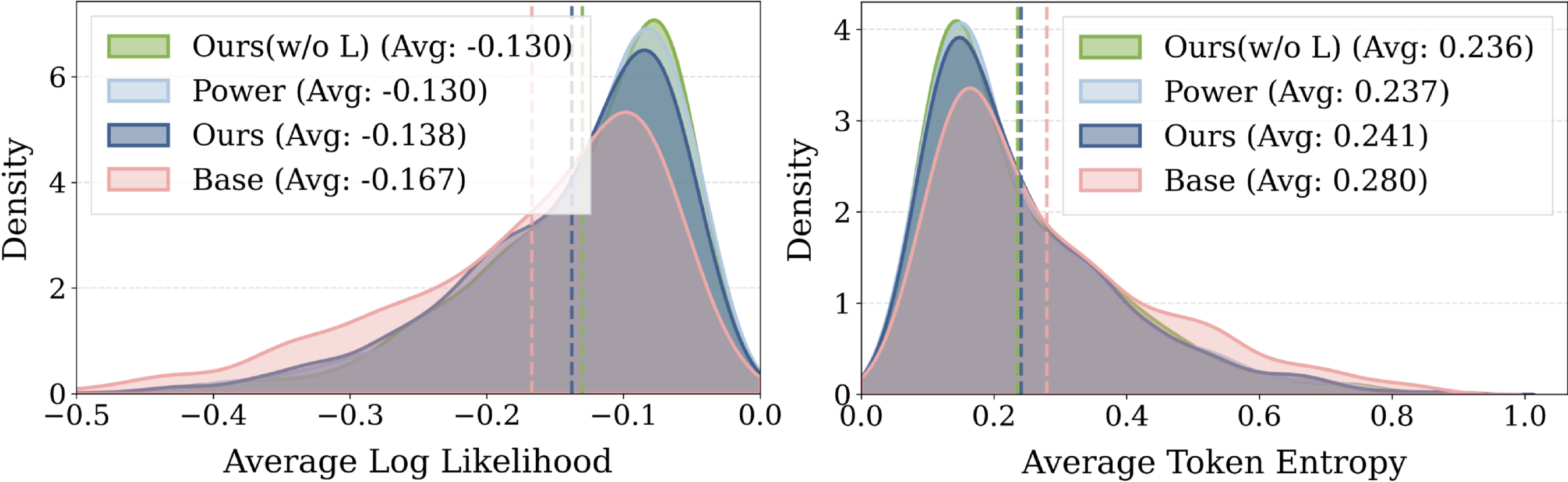}
  \caption{\textbf{Likelihood and entropy of Qwen2.5-VL-7B responses on MathVista, averaged by output length.}}
  \label{fig:like}
\end{figure}

\begin{table}[t]
\centering
\small
\caption{\textbf{Response length and token consumption on three datasets.} Multipliers (shown after $\times$) report token consumption relative to greedy decoding ($\times 1.0$).}

\label{tab:cost_stats}
\begin{tabular}{p{30mm}|p{13mm}|p{13mm}|p{13mm}}
\toprule
Method & MathVista & MMStar & MathVision \\
\midrule
\multicolumn{4}{c}{\textit{Response Length}} \\
\midrule
Greedy Decoding & 242 & 184 & 566 \\
Multi-Objective Sampling & 216 & 164 & 401 \\
TG Structured Sampling & 235 & 187 & 528 \\
\midrule
\multicolumn{4}{c}{\textit{Token Consumption}} \\
\midrule
Greedy Decoding & \numboxl{\hphantom{0,}242}\numboxl{\hphantom{0}\crossgreen{1.0}}
& \numboxl{\hphantom{0,}184}\numboxl{\hphantom{0}\crossgreen{1.0}}
& \numboxl{\hphantom{00,}566}~~~\numboxl{\hphantom{0}\crossgreen{1.0}}\\
Multi-Objective Sampling 
& \numboxl{5,665}\numboxl{\crossgreen{23.4}} 
& \numboxl{4,282}\numboxl{\crossgreen{23.3}} 
& \numboxl{12,840}~~~\numboxl{\crossgreen{22.3}}\\
TG Structured Sampling 
& \numboxl{3,710}\numboxl{\crossgreen{15.3}} 
& \numboxl{2,964}\numboxl{\crossgreen{16.2}} 
& \numboxl{\hphantom{0}9,904}~~~\numboxl{\crossgreen{17.5}}\\
\bottomrule
\end{tabular}
\end{table}

\subsection{Reasoning Trace Likelihood and Entropy}
\label{sec:like}

Figure~\ref{fig:like} shows the kernel density estimates of average token entropy and log-likelihood for Qwen2.5-VL-7B responses on MathVista under power-scaled sampling (Power), our full refinement target (Ours), our target without the linguistic term (Ours w/o L), and vanilla decoding (Base). By design, both power-scaled sampling and our method shift samples toward higher-likelihood, lower-entropy regions of the base model, favoring more confident continuations. ``Ours (w/o L)'' achieves the lowest entropy with likelihood comparable to power sampling, highlighting the effect of visual sharpening and entropy regularization. Compared with pure power sampling, ``Ours'' retains slightly higher entropy and slightly lower sequence likelihood, reflecting the intended multi-objective trade-off.

\begin{figure}[t]
  \includegraphics[width=\columnwidth]{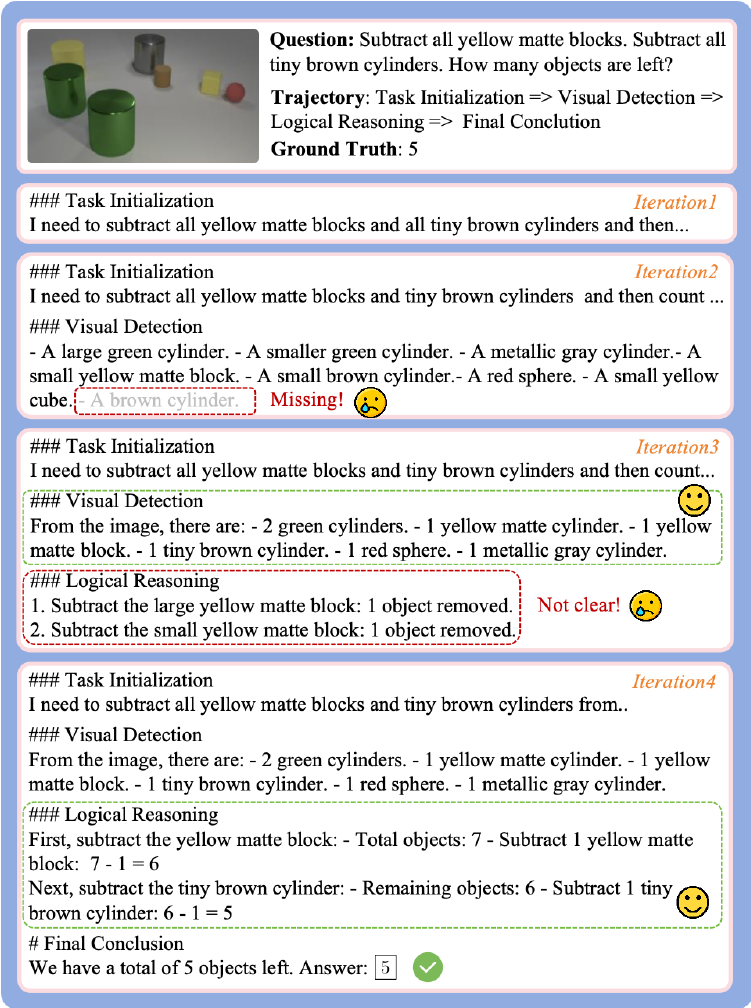}
  \caption{\textbf{Illustrative case study.}}
  \label{fig:case_study}
\end{figure}

\subsection{Output Length and Inference Cost}
\label{sec:efficiency}
\noindent\textbf{Response length.}
We do not observe a tendency toward longer responses under multi-objective sampling, as suggested
by \citet{karan2025reasoningsamplingbasemodel}. 
Instead, we observe a negative correlation where response length decreases as the alignment objective becomes stronger.
This trend indicates that enhanced alignment effectively suppresses linguistic redundancy and filler tokens inherent in vanilla decoding. Rather than relying on verbosity, our sampling objective directs the model toward the most discriminative reasoning steps, thereby increasing information density while preserving performance. 
By comparison, trajectory-guided structure sampling tends to produce longer responses than multi-objective sampling.

\noindent\textbf{Token consumption.} We define token consumption as the average number of output tokens generated per query. With visual calibration enabled, each output token incurs two forward passes, and we therefore count each token twice when computing token consumption. As shown in Table~\ref{tab:cost_stats}, sampling from the unnormalized target  distribution incurs substantial inference-time compute: multi-objective sampling increases token consumption by about $22$--$24\times$ over greedy decoding. In contrast, trajectory-guided structured 
sampling reduces token consumption to $15$--$18\times$ and achieves a $1.3$--$1.5\times$ reduction relative to multi-
objective sampling, while exhibiting longer response length.

\subsection{Case Study}
Figure~\ref{fig:case_study} visualizes the initial reasoning state at each refinement iteration. Since iteration $i$ inherits the accepted refinement from the previous cycle, consecutive panels reveal how the trajectory evolves through local updates. Iteration~1 establishes the task, while Iteration~2 extends the trajectory with Visual Detection but omits the target brown cylinder, leaving the object inventory incomplete. The initial state of Iteration~3 restores the missing object while preserving the earlier task context. However, its Logical Reasoning remains incomplete because the subtraction does not account for the brown cylinder. Iteration~4 retains the corrected detection and refines the reasoning into explicit steps, $7-1=6$ and $6-1=5$, ultimately yielding the correct answer. This illustrates the capacity of our MCMC-based refinement to incrementally 
correct local inconsistencies while preserving the overall reasoning context.

\section{Related Work}
\subsection{Test-Time Alignment Methods}
Test-time alignment adapts a frozen model to new objectives, distribution shifts, or evolving user preferences at inference time rather than updating parameters, encompassing a broad range of methods unified by post-hoc output steering.
First, hypothesis reweighting~\citep{lee2025inferencetime} leverages a pre-trained diverse ensemble of model heads and dynamically fits mixture weights at inference time using a small target-domain adaptation set.
Second, reward-guided decoding~\citep{xu2025genarmrewardguidedgeneration} achieves efficient test-time alignment by reshaping token-level probabilities with an autoregressive reward model. Recent advances~\citep{lin2025parm} further extend this paradigm by employing a single preference-conditioned model, which overcomes the inefficiency of model ensembling and allows different alignment criteria to be dynamically balanced on the fly. 
Third, control and planning approaches cast decoding as an inference-time optimization problem, using sampling-based control in pre-logit space~\citep{kanai2025testtimealignmentllmssamplingbased} or subgoal-guided predictive planning to maintain long-horizon consistency~\citep{wang2025testtimealignmentlargelanguage}.
Motivated by this view, we propose a sampling-based test-time alignment method that refines LVLM generations via trajectory-guided structured MCMC sampling.


\subsection{MCMC-based Autoregressive Sampling}
Several works have revisited classical Monte Carlo inference as a test-time alternative to training-time alignment, combining MCMC-style updates with autoregressive generation. 
For instance, \citet{zhaotsm2024} cast controlled generation as probabilistic inference and introduce twisted Sequential Monte Carlo, learning prefix-level twists to allocate particles toward high-reward continuations. 
Alternatively, \citet{faria2024quest} employ Metropolis--Hastings decoding to target a metric-induced Gibbs distribution via iterative accept-reject refinement with autoregressive proposals.
More recently, \citet{karan2025reasoningsamplingbasemodel} propose sampling from a power-scaled distribution defined directly by the base model itself. 
Our work is the first to investigate these principles for LVLMs. We introduce an efficient block-wise Metropolis--Hastings refinement steered by retrieved trajectories, significantly reducing the overhead of long-horizon generation. To ensure generation quality, we further enhance the sampling target with a synergistic objective.

\section{Conclusion}

This paper presents a sampling-based test-time alignment framework for LVLMs. Our approach is based on an automated, agent-driven trajectory learning algorithm to build a reasoning memory bank, where each trajectory captures high-level reasoning patterns for solving complex multimodal problems. To address the computational challenges of long-horizon generation, we propose trajectory-guided structured sampling, which enables localized refinement instead of costly full-sequence resampling. The retrieved guidance trajectory provides a global prior over reasoning structure and step order, making local refinement practical while preserving overall coherence. On top of this sampler, our synergistic alignment objectives steer generation toward visually grounded, reliable, and non-degenerate outputs. 

Experiments on five challenging datasets demonstrate that the framework achieves consistent performance gains without parameter updates, while substantially reducing token consumption relative to full-sequence resampling. Overall, our results position trajectory-guided test-time sampling as a practical and efficient paradigm for alignment in complex multimodal reasoning.

\begin{acks}
This work was supported by the New Generation Artificial Intelligence
National Science and Technology Major Project
(Grant No.~2025ZD0123402),
the Computational Biology Program
(Grant No.~25JS2830402)
of Science and Technology Commission of Shanghai Municipality
(STCSM),
and the Shanghai Municipal Science and Technology Major Project
(Grant No.~2025SHZDZX025G06).
\end{acks}

\bibliographystyle{ACM-Reference-Format}
\bibliography{sample-base}

\clearpage
\appendix
\section{Implementation Details}
\subsection{Definition of Reasoning Patterns}
\label{app:def_patterns}
\begin{table}[t]
\centering
\small
\caption{\textbf{Reasoning patterns with explicit interfaces.}}
\label{tab:patterns_io}
\begin{tabular}{p{24mm}|p{16mm}|p{30mm}}
\toprule
\textbf{Reasoning pattern} & \textbf{Inputs} & \textbf{Outputs} \\
\midrule
Task Initialization & Question $X$ & A single task  \\
\midrule
Decompose Problem &  A single task & A list of subtasks \\
\midrule
Visual Detection & A single task & A list of information (derive from image) \\
\midrule
Knowledge Retrieval & A single task & A list of information (derive from parametric knowledge) \\
\midrule
Logical Reasoning & A single task & A list of information (derive from deduction) \\
\midrule
Self-Verification & A list of information & A list of verified information \\
\midrule
Final Conclusion & A list of information & Final answer \\
\bottomrule
\end{tabular}
\end{table}

Table~\ref{tab:patterns_io} defines seven reasoning patterns as step functions with explicit input--output interfaces. Each step consumes either the question $X$, a single task, or an accumulated information list, and produces a task (or subtask list), an information list, or the final answer. Notably, while the table specifies the formal I/O signatures, the actual execution of each pattern also involves internal reasoning processes that transform inputs into outputs. This design turns free-form chain-of-thought into composable, well-structured trajectories, making complex behaviors such as decomposition and grounding easier to elicit and reuse.

Such pseudo-code-style definitions also support our downstream pipeline. They make trajectories comparable at the step level for retrieval and voting, enable sliding-window sampling by regenerating selected patterns while keeping others fixed, and simplify library curation through lightweight checks on structured outputs. Moreover, when scaling to more patterns, these interfaces make it straightforward to introduce rule-based behavior filters that accept, reject, or reroute pattern executions based on their inputs/outputs, steering reasoning toward desired behaviors. Overall, the pattern interfaces yield interpretable trajectories that are easier to construct and are more amenable to step-aligned refinement.

\subsection{Entropy-Weighted Multi-View Retrieval}
\label{app:retrieval}

For a query instance $(V, X)$, we initially retrieve a candidate pool of size $M{=}100$ via question-only embedding similarity. To account for the heterogeneous information across modalities, we propose an adaptive re-ranking mechanism based on entropy-weighted multi-view fusion. We consider three complementary perspectives: textual semantic similarity $s_{\text{text}}$, visual consistency $s_{\text{img}}$, and lexical $n$-gram overlap $s_{\text{ng}}$.
To dynamically modulate the importance of each view, we utilize an information-theoretic approach to quantify their respective utilities. Specifically, for each view $i$, we define the probability distribution $p_i$ over the candidate pool as:
\begin{equation}
p_i^{(j)}=\frac{s_i^{(j)}+\epsilon}{\sum_{j'=1}^{M}\left(s_i^{(j')}+\epsilon\right)}.
\end{equation}
The reliability of each modality is then inversely proportional to its normalized entropy $H_i$:
\begin{equation}
H_i=-\frac{1}{\log M}\sum_{j=1}^{M} p_i^{(j)}\log p_i^{(j)}.
\end{equation}
Intuitively, a view that yields a highly skewed distribution (low entropy) is more informative for ranking than one with a uniform distribution (high entropy). The final retrieval score $S^{(j)}$ is computed as a weighted sum $\sum_i w_i s_i^{(j)}$, where $w_i \propto (1-H_i)$. This ensures that the top-$k$ results are selected based on the most reliable and discriminative signals for the given query. We use Qwen3-Embedding-0.6B as our text embedding model and clip-vit-base-patch32 as our image embedding model.

\subsection{Design of Difficulty Estimator}
\label{app:difficulty}

After obtaining the guidance trajectory, we apply a simple rule-based estimator to route instances by difficulty. We first remove behaviors that do not directly advance the reasoning process, including \texttt{Task Initialization}, \texttt{Self-Verification}, and \texttt{Final Conclusion}. If the remaining trajectory contains more than one action, we treat the instance as non-trivial, suggesting that it likely requires multi-hop reasoning; we then perform block-wise sampling under trajectory guidance. Otherwise, we treat it as trivial and apply full-sequence MCMC without trajectory guidance, which avoids overthinking while improving inference efficiency.


\subsection{MCMC for Full-Sequence Sampling}
\label{app:MCMC}

For full-sequence sampling, we employ the autoregressive MH sampler proposed by \citet{karan2025reasoningsamplingbasemodel}. To ensure sampling efficiency within high-dimensional spaces, a sequence of maximum length $T$ is partitioned into $K$ blocks, each consisting of $T/K$ tokens. The refinement process follows $K$ sequential cycles, during which the number of active blocks eligible for resampling and MH updates increases incrementally. The procedure terminates early if an end-of-sequence (EOS) token is generated. In our implementation, we set the maximum sequence length $T=3072$ and the number of blocks $K=16$.

\section{More Experiments}
\subsection{Generation Entropy and Answer Accuracy}
\label{app:evsacc}
\begin{figure}[t]
  \includegraphics[width=\columnwidth]{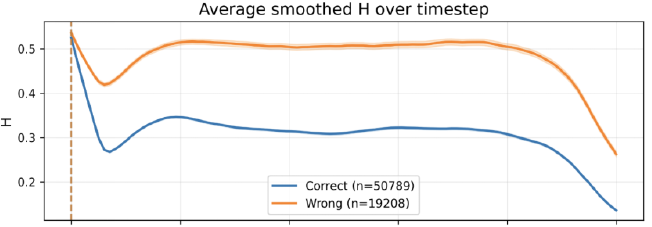}
  \caption{\textbf{Average smoothed entropy over timestep.}}
  \label{fig:entropy_acc}
\end{figure}

In Figure~\ref{fig:entropy_acc}, we plot the average smoothed token-level output entropy $H$ over decoding trajectories on the ThinkLite-70k dataset. A consistent gap is observed throughout generation: trajectories that
lead to correct answers exhibit markedly lower entropy than incorrect ones. While this trend suggests that entropy can serve as a proximal reliability signal for test-time refinement, it is not an infallible metric. The model may occasionally succumb to being confidently wrong, yielding low-entropy but factually incorrect outputs that stifle beneficial exploration. To mitigate this, our framework employs entropy as a regularization term coupled with a vision-aware calibration mechanism. This dual approach effectively filters out low-entropy hallucinations, leveraging entropy to stabilize the refinement process rather than as a binary criterion for correctness.



\subsection{Analysis on Response Length}
\label{app:length}
\begin{table}[t]

\centering
\small
\caption{\textbf{Response length of different sampling
objectives on MathVista and MMStar with Qwen2.5-VL-7B.}}
\label{tab:length_stats}
\begin{tabular}{p{48mm}|p{13mm}|p{13mm}}
\toprule
Method & MathVista & MMStar  \\
\midrule

Greedy Decoding & 242 & 184  \\
Power Sampling & 229 & 171  \\
Power Sampling + Visual + Entropy & 218 & 164  \\
\bottomrule
\end{tabular}
\end{table}

\begin{figure}[t]
  \includegraphics[width=\columnwidth]{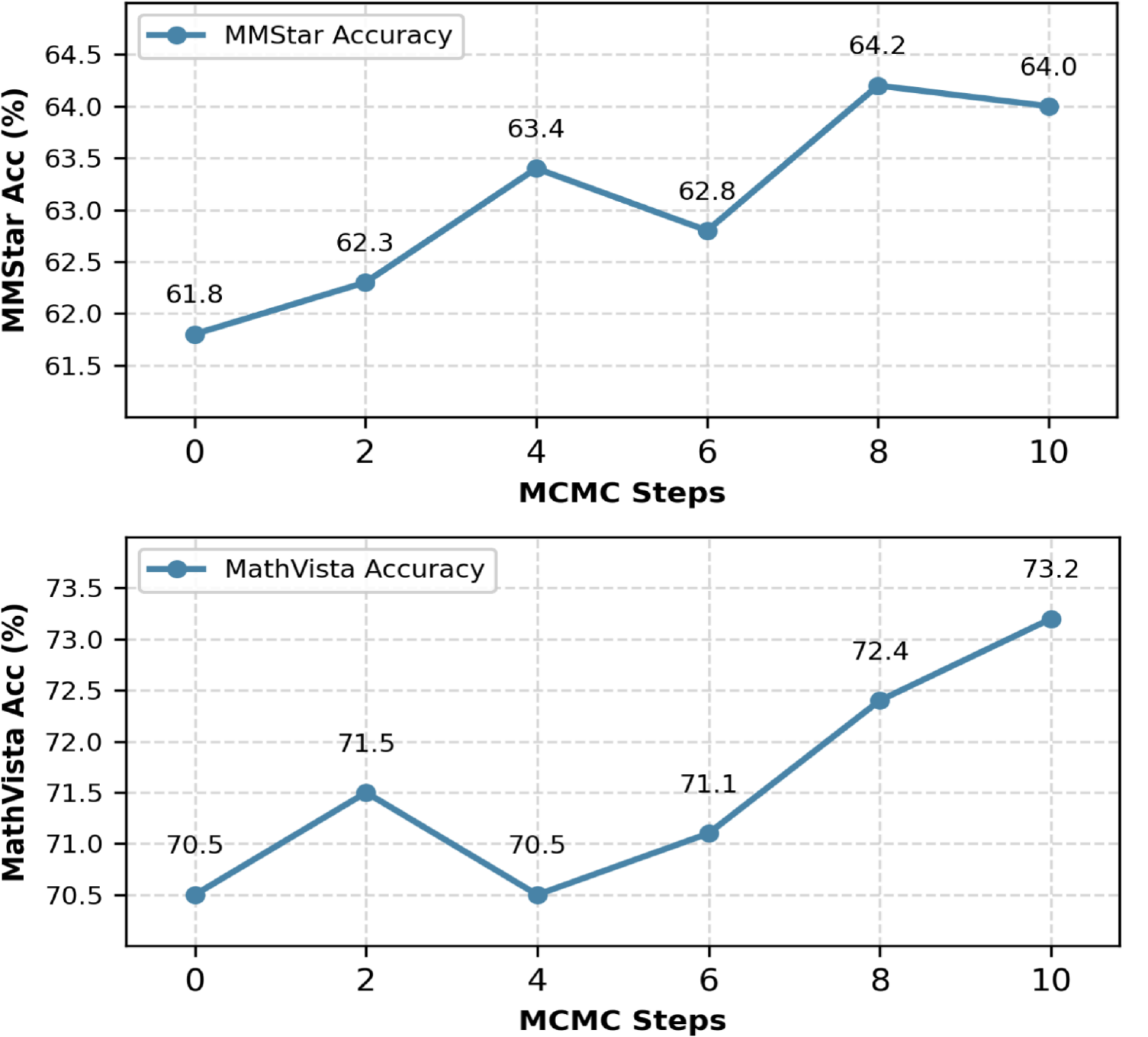}
  \caption{\textbf{Effect of MCMC refinement steps on MathVista and MMStar with Qwen2.5-VL-7B.}}
  \label{fig:mcmcstps}
\end{figure}

\begin{figure*}[t]
  \includegraphics[width=\textwidth]{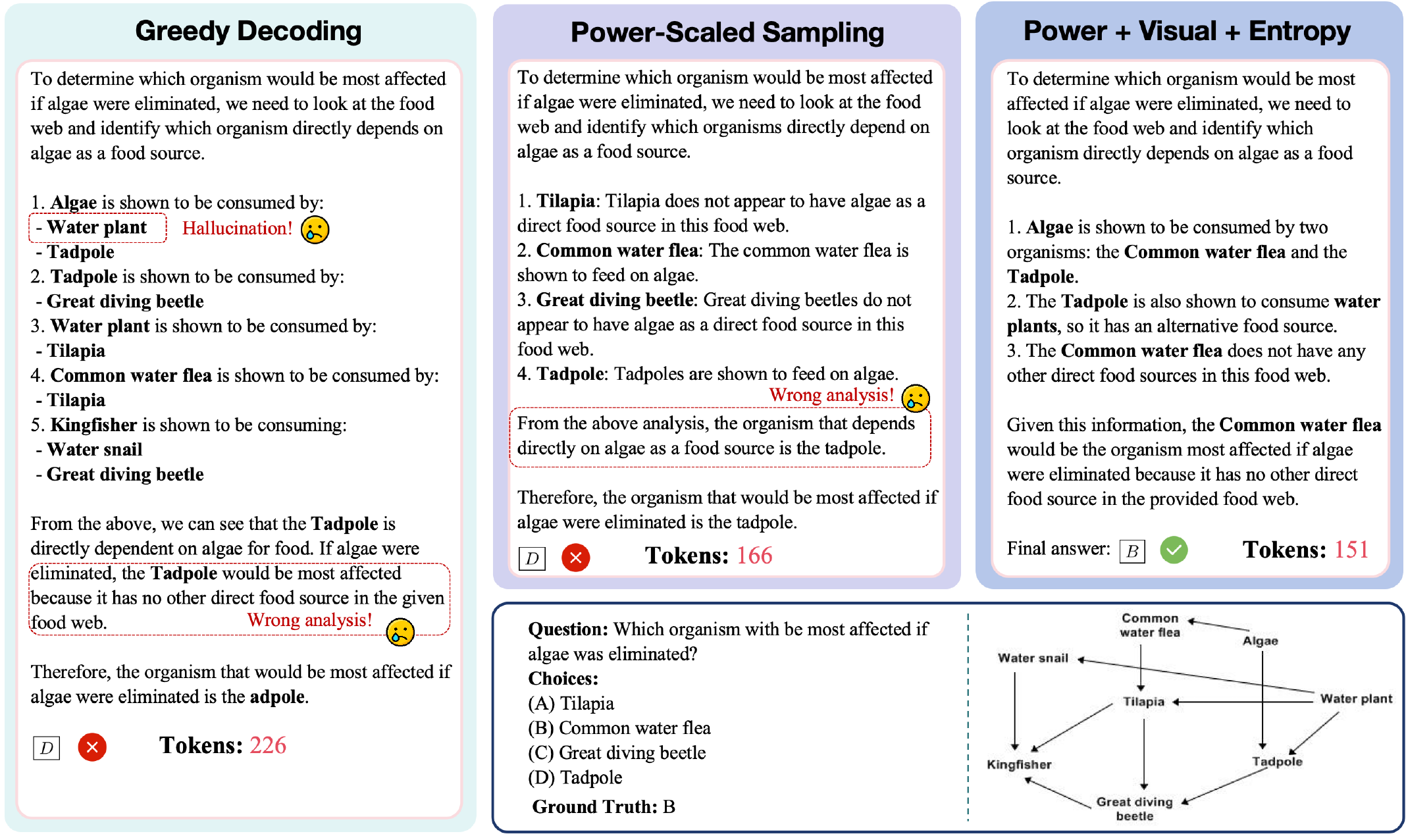}
  \caption{\textbf{Case study of different sampling variants.}}
  \label{fig:length_case}
\end{figure*}
Table~\ref{tab:length_stats} reports the average response length across three sampling configurations. We observe a consistent trend where the introduction of additional alignment objectives leads to a reduction in average response length.
First, naive power sampling leads to shorter responses, which contrasts with the findings of \citet{karan2025reasoningsamplingbasemodel}. We attribute this to the tendency of LVLMs to lose attention to visual inputs as generation length increases \citep{xu2025more,tian2025thoughtaccuracydualnature}, leading to performance degradation. We leave a detailed study for future work.
Vanilla decoding is frequently bottlenecked by linguistic redundancy, where high-probability but uninformative sequences dominate the output. By coupling visual evidence with entropy-based weighting, our method actively penalizes such filler tokens, redirecting the generation toward critical, discriminative reasoning steps. As a result, the refined trajectories deliver higher accuracy with remarkable conciseness, fundamentally enhancing the information density of the underlying reasoning. A qualitative case study is provided in Figure~\ref{fig:length_case}. Specifically, greedy decoding and power-scaled sampling tend to generate either irrelevant contextual fillers or hallucinatory descriptions that contradict the visual evidence, ultimately culminating in incorrect conclusions. In contrast, by filtering out these linguistically dominant but visually inconsistent paths, our method directs the model to converge on the correct answer with fewer tokens. This case illustrates that our approach effectively prunes redundant or divergent reasoning branches, ensuring that the generated trajectory remains both concise and strictly grounded in the provided image.

\subsection{Scaling Law on MCMC Steps}
\label{sec:scaling_mcmc}

$N_{\text{MCMC}}$ controls test-time scaling in our framework by determining the number of Metropolis--Hastings refinement steps performed under the alignment target. Each step corresponds to one propose--accept update, so increasing $N_{\text{MCMC}}$ allocates more decoding-time compute to sequence-level correction, while smaller $N_{\text{MCMC}}$ favors lower latency. Figure~\ref{fig:mcmcstps} shows that increasing $N_{\text{MCMC}}$ generally improves performance on both MMStar and MathVista, although the gains are not strictly monotonic at intermediate steps. On MMStar, accuracy rises from 61.8 at $N_{\text{MCMC}}=0$ to 64.2 at $N_{\text{MCMC}}=8$, with a slight drop to 64.0 at $N_{\text{MCMC}}=10$, suggesting that most of the benefit is already realized within a moderate refinement budget. On MathVista, the trend is more variable at small step counts but becomes clearly positive at larger budgets, improving from 70.5 at $N_{\text{MCMC}}=0$ to 73.2 at $N_{\text{MCMC}}=10$. Overall, these results suggest a favorable test-time scaling trend with respect to $N_{\text{MCMC}}$: allocating more MCMC refinement steps generally improves reasoning accuracy, although the returns become less smooth and dataset-dependent at intermediate budgets.

\subsection{Evaluation on Hallucination Benchmarks}

\label{app:hallu}
\begin{table}[t]

\centering
\small
\caption{\textbf{Evaluation of hallucination on Qwen2.5-VL-7B. We report the $F_1$ score for POPE and CHAIR scores (lower scores indicate fewer hallucinations.
).}} 
\label{tab:hallu}
\begin{tabular}{p{34mm}|p{12mm}|p{12mm}|p{12mm}}
\toprule
Method
& POPE $F_1 \uparrow$
& CHAIR$_S \downarrow$
& CHAIR$_I \downarrow$ \\
\midrule

Greedy Decoding & 85.9 & 38.8 & \textbf{9.2} \\
Power Sampling & 85.8 & 39.2 & 9.3 \\
Power Sampling + Visual & \textbf{86.1} & \textbf{38.6} & \textbf{9.2} \\
\bottomrule
\end{tabular}
\end{table}
To evaluate the effectiveness of vision-aware distribution sharpening in mitigating hallucinations, we perform a quantitative analysis across two standard benchmarks, as summarized in Table \ref{tab:hallu}.
We evaluate object hallucinations using POPE \cite{li-etal-2023-evaluating}, a VQA-based probing protocol, and CHAIR \cite{rohrbach-etal-2018-object}, which quantifies hallucinations in image captioning by cross-referencing generated tokens with ground-truth objects. As formulated in Equation \eqref{eq:chair_i} and \eqref{eq:chair_s}, CHAIR measures hallucination at both the instance ($\text{CHAIR}_I$) and sentence ($\text{CHAIR}_S$) levels:

\begin{equation}
\label{eq:chair_i}
\text{CHAIR}_I = \frac{|\{\text{hallucinated objects}\}|}{\text{all mentioned objects}}, 
\end{equation}

\begin{equation}
\label{eq:chair_s}
\text{CHAIR}_S = \frac{|\{\text{captions with hallucinated objects}\}|}{\text{all captions}}.
\end{equation}

As reported in Table \ref{tab:hallu}, power sampling with visual calibration yields marginal yet consistent improvements across the evaluated benchmarks. Specifically, our method achieves a POPE $F_1$ score of 86.1, slightly surpassing the greedy decoding baseline. Notably, while vanilla power sampling exhibits a slight increase in hallucination rates (e.g., 39.2 $\text{CHAIR}_S$), the integration of visual priors successfully mitigates this regression, reaching the lowest $\text{CHAIR}_S$ of 38.6. These results suggest that our approach helps maintain factual alignment with the visual input.

\subsection{Hyperparameter sensitivity}
\begin{figure}[t]
  \includegraphics[width=\columnwidth]{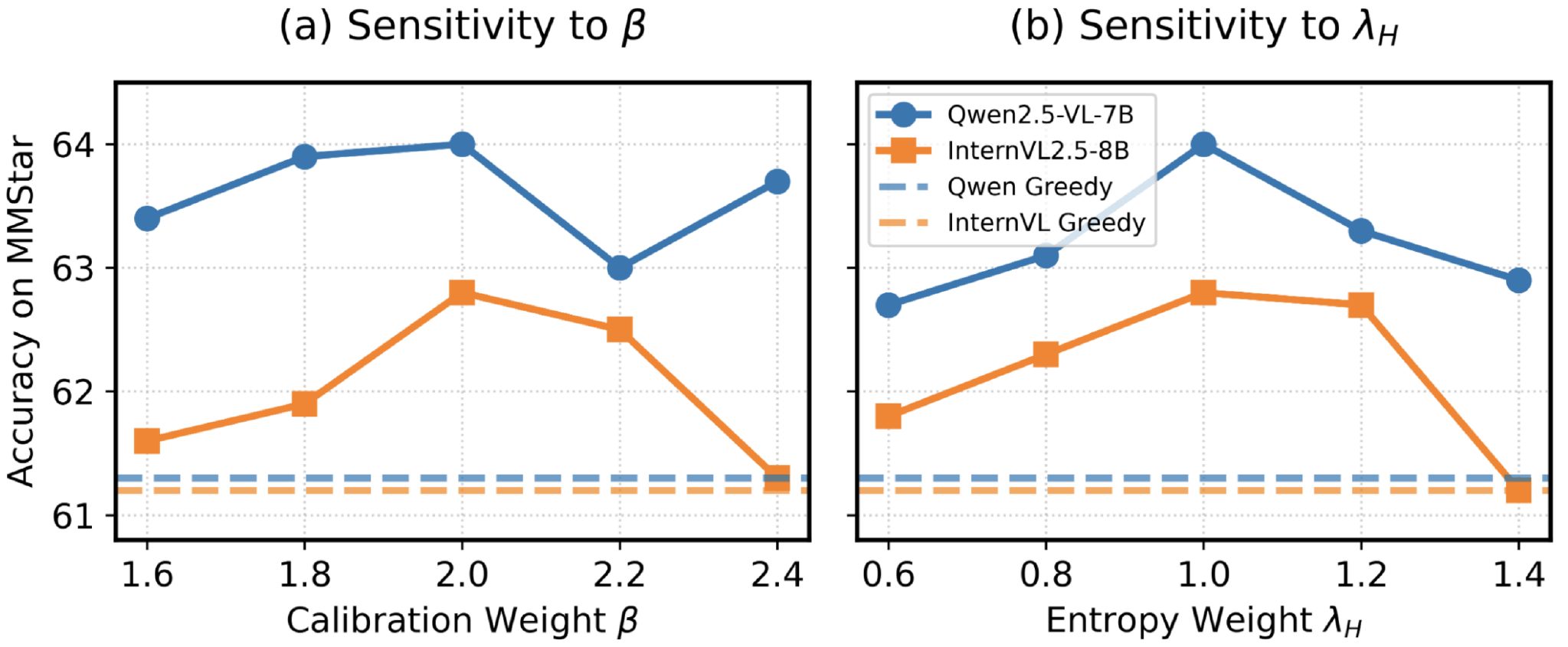}
  \caption{Hyperparameter sensitivity analysis on the MMStar dataset with Qwen2.5-VL-7B and InternVL2.5-8B.
}
  \label{fig:sensi}
\end{figure}
Figure~\ref{fig:sensi} examines the sensitivity of Qwen2.5-VL-7B and InternVL2.5-8B to the calibration weight $\beta$ and entropy weight $\lambda_H$ on MMStar. Both models attain their highest accuracy at approximately $\beta=2.0$ and $\lambda_H=1.0$. Performance remains relatively stable under moderate deviations from these values, indicating that the method is not overly sensitive to either hyperparameter, although more extreme settings cause noticeable degradation. Importantly, all evaluated configurations consistently outperform their corresponding greedy baselines, demonstrating the robustness of the proposed method across a broad range of hyperparameter choices.

\section{Visualization}
\begin{figure}[t]
  \includegraphics[width=\columnwidth]{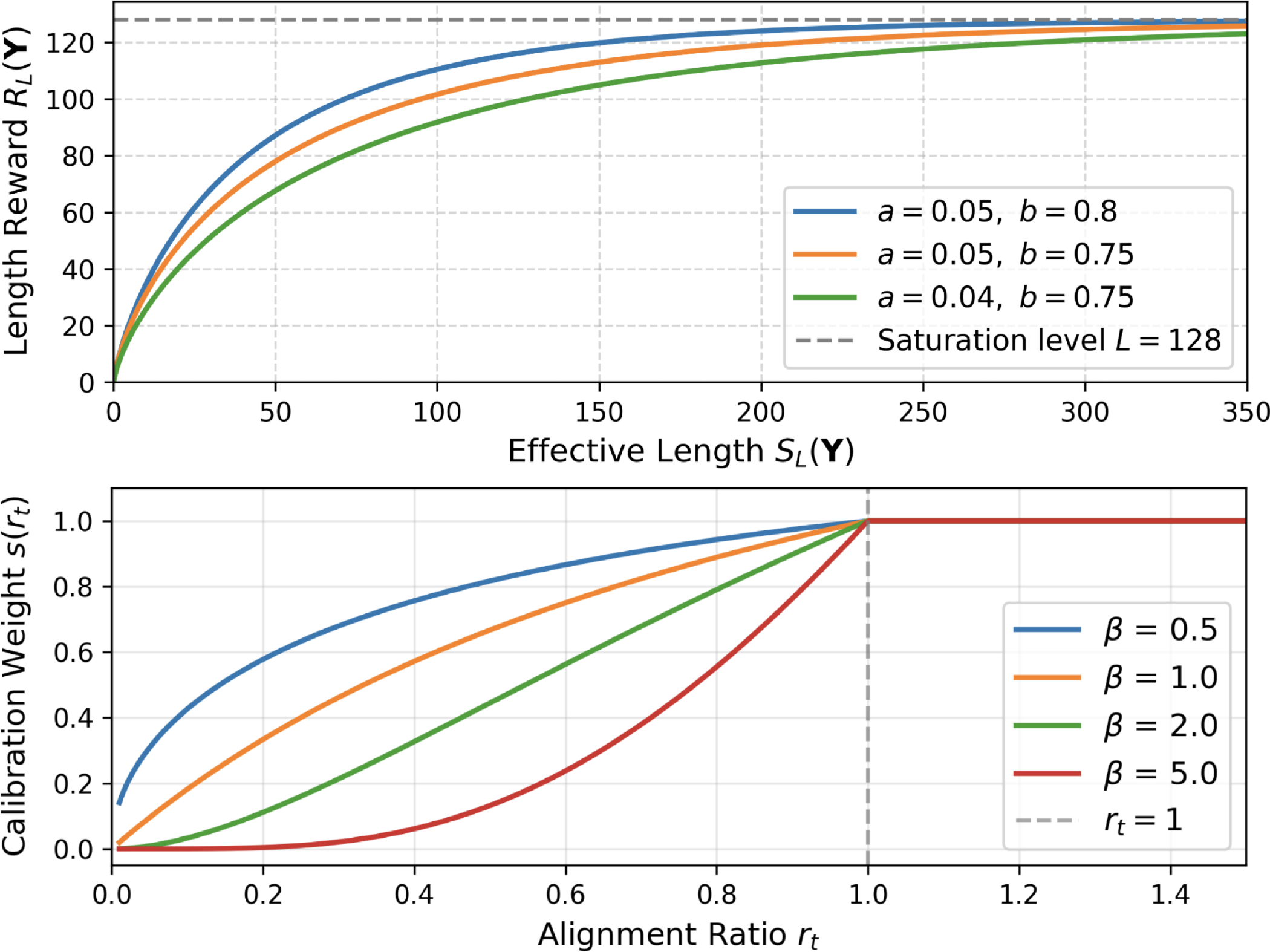}
  \caption{\textbf{Visualization of vision-aware calibration weight $s(r_t)$ and length reward $R_L(\mathbf{Y})$}}
  \label{fig:calibration}
\end{figure}

Figure~\ref{fig:calibration} visualizes the two shaping functions used in our alignment target. The top panel shows the effective-length reward $R_L(\mathbf{Y})$ as a function of $S_L(\mathbf{Y})$. In all cases, the reward increases monotonically with effective length and gradually saturates at the target level $L$, encouraging sufficiently long reasoning traces while avoiding unbounded growth. The parameters $a$ and $b$ control the growth rate and saturation behavior: larger values lead to faster early growth and earlier saturation, while smaller values produce a smoother increase. In our experiments, we set $a=0.05$, $b=0.8$ for all settings.

The bottom panel shows the vision-aware calibration weight $s(r_t)$ as a function of the alignment ratio $r_t$. The weight is bounded in $[0,1]$, increases monotonically with $r_t$, and reaches $1$ when $r_t \ge 1$, so tokens that are better supported by visual evidence receive full sharpening. The parameter $\beta$ controls the selectivity of this calibration: smaller $\beta$ yields a smoother transition, whereas larger $\beta$ suppresses weakly aligned tokens more aggressively. We use $\beta = 2.0$ in all experimental settings.

\end{document}